\pdfoutput=1

\documentclass[11pt]{article}

\usepackage{ACL2023}

\usepackage{times}
\usepackage{svg}
\usepackage{latexsym}
\usepackage{amsfonts}
\usepackage{amsmath}
\usepackage{float}
\usepackage{booktabs}
\usepackage[T1]{fontenc}

\usepackage[utf8]{inputenc}

\usepackage{microtype}

\usepackage{inconsolata}

\usepackage{CJKutf8}
\usepackage{graphicx}
\usepackage{hyperref}
\usepackage{dsfont}
\title{Why Does Zero-Shot Cross-Lingual Generation Fail?\\ An Explanation and a Solution}

\author{Tianjian Li$^{1}$ \and Kenton Murray$^{1,2}$ \\
    $^{1}$Center for Language and Speech Processing \\
    $^{2}$Human Language Technology Center of Excellence \\
    Johns Hopkins University \\
    \{\texttt{tli104, kenton}\}@\texttt{jhu.edu}}

\begin{document}
\maketitle
\begin{abstract}
Zero-shot cross-lingual transfer is when a multilingual model is trained to perform a task in one language and then is applied to another language. 
Although the zero-shot cross-lingual transfer approach has achieved success in various classification tasks \citep{wu-dredze-2019-beto}, its performance on natural language generation tasks falls short in quality (\citealp{ronnqvist-etal-2019-multilingual}; \citealp{vu-2022-overcoming}) and sometimes outputs an incorrect language \citep{xue-etal-2021-mt5}. In our study, we show that the fine-tuning process learns language invariant representations, which is beneficial for classification tasks but harmful for generation tasks. Motivated by this, we propose a simple method to regularize the model from learning language invariant representations and a method to select model checkpoints without a development set in the target language, both resulting in better generation quality. Experiments on three semantically diverse generation tasks show that our method reduces the accidental translation problem by 68\% and improves the ROUGE-L score \citep{lin-2004-rouge} by 1.5 on average.


\end{abstract}

\section{Introduction}

Language Models (LMs) pre-trained on multilingual corpora (\citealp{devlin-etal-2019-bert}; \citealp{conneau-etal-2020-unsupervised}; \citealp{liu-etal-2020-multilingual-denoising}; \citealp{xue-etal-2021-mt5}) exhibit zero-shot cross-lingual transfer ability \citep{wu-dredze-2019-beto}. Given only annotated data in one language for a task,  multilingual LMs are able to perform this task in languages seen only during the pre-training stage. The cross-lingual transferability of multilingual LMs reduces the need for annotated data in low-resource languages, which is valuable for building practical multilingual NLP systems. 

Existing studies on cross-lingual transfer select tasks such as word alignment \citep{artetxe-etal-2020-cross}, POS tagging \citep{pires-etal-2019-multilingual}, dependency parsing and sentence classification \citep{wu-dredze-2019-beto} to investigate cross-lingual transferability of multilingual LMs \citep{pmlr-v119-hu20b}, and few works focus on cross-lingual transfer in generation tasks (\citealp{maurya-etal-2021-zmbart}; \citealp{maurya-desarkar-2022-meta}). Cross-lingual transfer approach in generation tasks are known to produce incoherent text \citep{ronnqvist-etal-2021-multilingual}, generate in a wrong language \citep{xue-etal-2021-mt5}, and suffer from catastrophic forgetting \citep{vu-2022-overcoming}. Table \ref{examples} illustrates a common problem where the multilingual LM generates text in an incorrect language. Moreover, such a problem becomes more severe when under a true zero-shot setting (\citealp{zhao-etal-2021-closer}; \citealp{dontstop-emnlp-2022}), where we do not have annotated data in the target language to guide model selection.

We show that the reason why zero-shot cross-lingual transfer in text generation fails is because the \textbf{fine-tuning process learns language invariant representations}, which is beneficial for classification tasks, but detrimental to generation tasks. In our paper, we use the cosine similarity between parallel sentence representations in different languages to measure the Cross-Lingual Representation Similarity (\textbf{XLRS}). We use a range of tasks from classification to extractive question answering, then to abstractive generation to show that in the best performing model, the XLRS after fine-tuning decreases as we move from classification to generation.

\begin{table*}[]
\centering
\begin{tabular}{@{}ll@{}}
\toprule
Prediction & Speak to your doctor, understand the dangers of alcohol consumption..                                 \\
Target     & \begin{CJK*}{UTF8}{gbsn} 与您的医生交谈：改变您对戒酒的想法。 \end{CJK*}                  \\
           & (Speak to your doctor: change your thinking towards quitting alcohol)           \\ \midrule
Prediction & Review accounting books periodically.                                           \\
Target     & \begin{CJK}{UTF8}{} \CJKfamily{mj} 기간을 결정한다. 회계장부를 모두 검토한다. 누락된 정보를 취합한다.   \end{CJK}                                        \\
           & (Determine the period. Review all accounting books. Gather missing information.) \\ \bottomrule
\end{tabular}
\caption{Example predictions of mT5 model fine-tuned on English WikiHow instructions and evaluated on Chinese and Korean input. The model outputs relevant text in an incorrect language.}
\label{examples}
\end{table*}

The fact that language invariant representations causes the degradation in generation tasks challenges the common belief that invariant representations generally enhance cross-lingual transfer on all downstream tasks (\citealp{Cao2020Multilingual}; \citealp{conneau-etal-2020-emerging}; \citealp{yang2022enhancing}; \citealp{xian2022crosslingual}). To the best of our knowledge, our work is the first to provide an analysis of how XLRS affects cross-lingual transfer in language generation tasks. 


Motivated by our findings, we propose to use an auxiliary source language that implicitly regularizes the XLRS being too large and results in better generation performance over three complex natural language generation tasks (Summarization, Story Completion, and Title Generation). Under a true zero-shot setting,  choosing the model checkpoint with the lowest XLRS results in an average of 4.1 point increase in ROUGE-L over using a source development set in two generation datasets. 

To sum up, our contributions are threefold:
\begin{itemize}
    \item We show that fine-tuning on a single source language increases the cosine similarity between sentence representations of different languages (XLRS). 

    \item We show that the increase in XLRS causes degradation of cross-lingual transfer in generation tasks, and argue that the prevalent understanding of the benefit of similar representations does not apply to generation tasks. 
    \item  
    We empirically show that using two gold-annotated source languages instead of one
    regularizes the XLRS, resulting in an average increase of 1.5 in ROUGE-L.
\end{itemize}

\section{Related Works}
\textbf{Multilingual Language Models.} 
One line of work is to train multilingual versions of modern Language Models. \textbf{mBERT
} \citep{mBERT} is the multilingual version of BERT \citep{devlin-etal-2019-bert}, which uses the same encoder-only model architecture but is only trained on multilingual corpora. \textbf{XLM-R} \citep{conneau-etal-2020-unsupervised} is the multilingual version of RoBERTa \citep{liu2019roberta}, which implements multiple optimization tricks and is larger in scale, resulting in better performance than BERT. \textbf{mBART} \citep{liu-etal-2020-multilingual-denoising} is the multilingual version of BART \citep{lewis-etal-2020-bart}, an encoder-decoder model trained to reconstruct the original text through various types of artificially introduced noises. 
\textbf{mT5} \citep{xue-etal-2021-mt5} is the multilingual version of T5 \citep{2020t5}, an encoder-decoder model trained on a span denoising objective. 

\noindent \textbf{Cross-lingual Transfer.} Multilingual models are able to be fine-tuned on annotated data of a task in only one source language and transfer the knowledge to other target languages to perform the same task without any supervision. While \citet{pires-etal-2019-multilingual} states that sub-word overlap between source and target facilitates cross-lingual transfer, \citet{K2020Cross-Lingual} shows that cross-lingual transfer manifests in pairs of source and target with zero sub-word overlap and word order is instead the most crucial ingredient. The performance of cross-lingual transfer between languages with a different order severely drops. Although the importance of word order is echoed by later studies (\citealp{artetxe-etal-2020-cross}; \citealp{dufter-schutze-2020-identifying}), recent studies have also debated in favor of the importance of matching script also contributing to cross-lingual transfer (\citealp{lauscher-etal-2020-zero}; \citealp{fujinuma-etal-2022-match}). 
\citet{wu-etal-2022-zero} points out that the optimal set of parameters that generalizes well to all languages is a subset of parameters that achieves good performance on the source language. Therefore it is hard to find the optimal zero-shot cross-lingual transfer parameters by only optimizing source language performance. \citet{chen-ritter-2021-model} train a scoring model with the input features being the model's hidden representations and the output score being how well it generalizes to a given target language. However, previous studies focus on lower-level NLP tasks, which include text classification, dependency parsing, and extractive question answering \citep{pmlr-v119-hu20b} and rarely touch on language generation. \\
\indent Another line of work focuses on applying cross-lingual transfer to a wide range of multilingual NLP applications, which include sequence tagging \citep{zhilin2016multitask}, Named Entity Recognition \citep{xie-etal-2018-neural}, dependency parsing \citep{ahmad-etal-2019-difficulties}, sentence classification (\citealp{conneau-etal-2018-xnli}; \citealp{yang-etal-2019-paws}), and information retrieval \citep{izacard2022unsupervised}. Empirical studies also train ranking models \citep{lin-etal-2019-choosing}, use meta-learning \citep{nooralahzadeh-etal-2020-zero}, or use Shapley Value \citep{parvez-chang-2021-evaluating} to predict which sources perform the best for a given target language. \\

\noindent \textbf{Natural Language Generation.} Multilingual LMs are prone to produce text that is repetitive \citep{xu2022learning}, contains hallucinations \citep{raunak-etal-2021-curious}, or is in the wrong language (\citealp{zhang-etal-2020-improving}; \citealp{xue-etal-2021-mt5}; \citealp{vu-2022-overcoming}). \citealp{vu-2022-overcoming} proposed to use parameter efficient fine-tuning methods (\citealp{lester-etal-2021-power}; \citealp{qin-eisner-2021-learning}; \citealp{li-liang-2021-prefix}) to regularize the model to generate in a desired language. Other ways to improve generation quality include using back translation (\citealp{gu-etal-2019-improved}; \citealp{zhang-etal-2020-improving}), and transliteration \citep{sun-etal-2022-alternative} as data augmentation techniques, mixing in the pretrain objective during fine-tuning \citep{xue-etal-2021-mt5} and using an auxiliary source language in machine translation \citep{xu-etal-2021-improving-multilingual}. Two concurrent efforts are close to our work: \citet{xu-murray-2022-por} and \citet{dontstop-emnlp-2022} both empirically show that using multiple languages during fine-tuning in few-shot cross-lingual transfer improves performance in text classification. Our work differs in that we evaluated \textbf{text generation} under a \textbf{true zero-shot setting}, where we have access to neither a few examples to train on nor an annotated development set to guide model checkpoint selection.

\section{Setup}
The consensus of the literature (\citealp{Cao2020Multilingual}; \citealp{conneau-etal-2020-emerging}; \citealp{tiyajamorn-etal-2021-language}; \citealp{yang2022enhancing}; \citealp{xian2022crosslingual}) is that if a model can produce similar representations for parallel sentences, the model would be able to achieve good cross-lingual transfer performance. Intuitively, if a model maps parallel sentences in English and French into nearly identical representations, and is able to predict the sentiment of the English sentence, it will also be able to predict the sentiment of the French sentence.

We hypothesize that the fine-tuning process increases the similarity between sentence representations of different languages. We use the following setups and tasks to verify our hypothesis.
\subsection{Models and Datasets}
\textbf{Models} We select the state-of-the-art multilingual language model: mT5-base \citep{xue-etal-2021-mt5}.
We use the Huggingface \citep{wolf-etal-2020-transformers} implementation. We use a uniform learning rate of 7e-5, a batch size of 32 for 10 epochs for all tasks described below.
\begin{table}[h]
\centering
\small
\begin{tabular}{@{}ccc@{}}
\toprule
Name       & Task                      & Metric  \\ \midrule
UDPOS        & Part-of-speech tagging    & Acc.  \\
PAWS-X     & Paraphrase Identification & F1  \\
TyDiQA     & Question Answering        & F1/EM  \\
WikiLingua & Summarization             & ROUGE \\ \bottomrule
\end{tabular}
\caption{Summary of tasks used in \S 5.}
\label{tasks}
\end{table}

\noindent \textbf{Datasets} 
Table \ref{tasks} describes the tasks we used in the following section to show the transition from classification to generation\footnote{We follow \citep{vu-2022-overcoming} and report the SP-ROUGE score.}. We use the \textbf{UDPOS} \citep{nivre2018universal} dataset containing sentences and the part-of-speech tag of each word. For sentence-level classification, we use the \textbf{PAWS-X} \citep{yang-etal-2019-paws} dataset containing pairs of sentences and a binary tag on whether the second sentence entails the first sentence. For extractive generation, we use the \textbf{TyDiQA-GoldP} \citep{Clark2020tydiqa} dataset which contains paragraphs and questions whose answers are spans extracted from the paragraphs. For abstractive generation, we use the \textbf{WikiLingua} \citep{ladhak-etal-2020-wikilingua} dataset, which contains WikiHow instructions and their summaries in 17 languages.  

We use the story completion (\textbf{SG}) and title generation (\textbf{TG}) task in the MTG benchmark \citep{chen-etal-2022-mtg}, a recently introduced benchmark to evaluate multilingual text generation. We follow \citealp{vu-2022-overcoming}, which uses the WikiLingua dataset to construct \textit{WikiLingua-0}, where the model is only fine-tuned on English and evaluated on other languages. We extend \textit{WikiLingua-0} and use languages besides English as the source and evaluate the zero-shot directions. 

In all of our experiments, we report the results averaged across three runs with different random seeds. For each source language, we only use the top 10k training examples to train our model to ablate the effect of training data size on cross-lingual transfer. Unless specified otherwise, we evaluate under a \textbf{true zero-shot} setting, where we select the model checkpoint based on its performance on a dev set of the source language. 

\subsection{Sequence to Sequence Learning}
We cast sequence labeling and sentence classification tasks into a text-to-text format using templates described in Table \ref{table1} in the Appendix. We follow \citet{karthik-seq2seq-2022} and cast sequence labeling tasks into the sentinel + tag format. We follow \citet{schick-schutze-2021-exploiting} and cast the sentence entailment task into a cloze question, supervising the model to predict the word "yes" for entailment and the word "no" for non-entailment.

\section{Learning Dynamics of Cross-lingual Transfer}
\begin{figure}[]
    \centering
    \includegraphics[width=180pt, height=110pt]{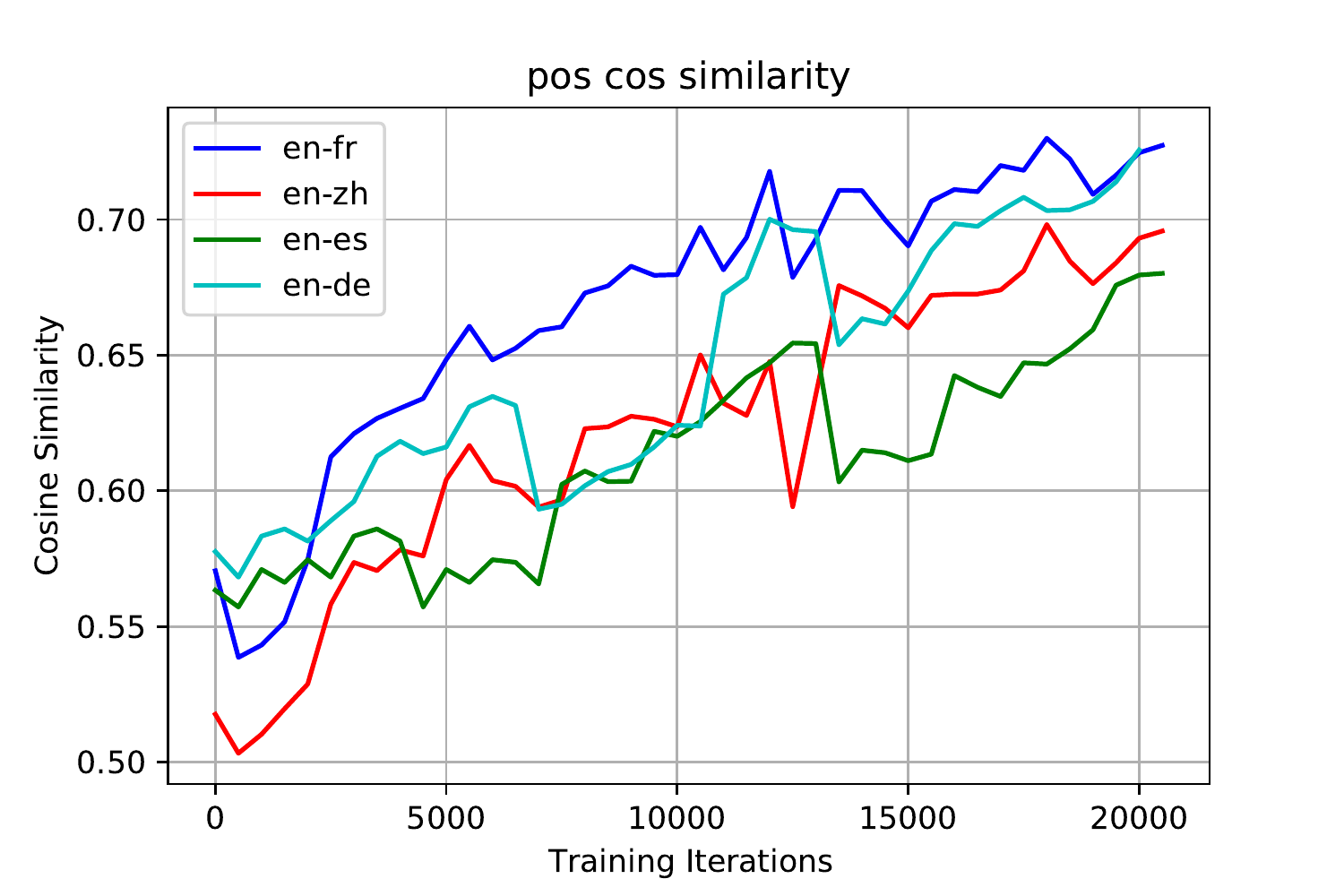}
    \caption{Average cosine similarity between parallel sentence representations (XLRS) in  pretrained mT5-base model fine-tuned on English POS tagging data.}
    \label{pos-cos}

    \includegraphics[width=180pt, height=110pt]{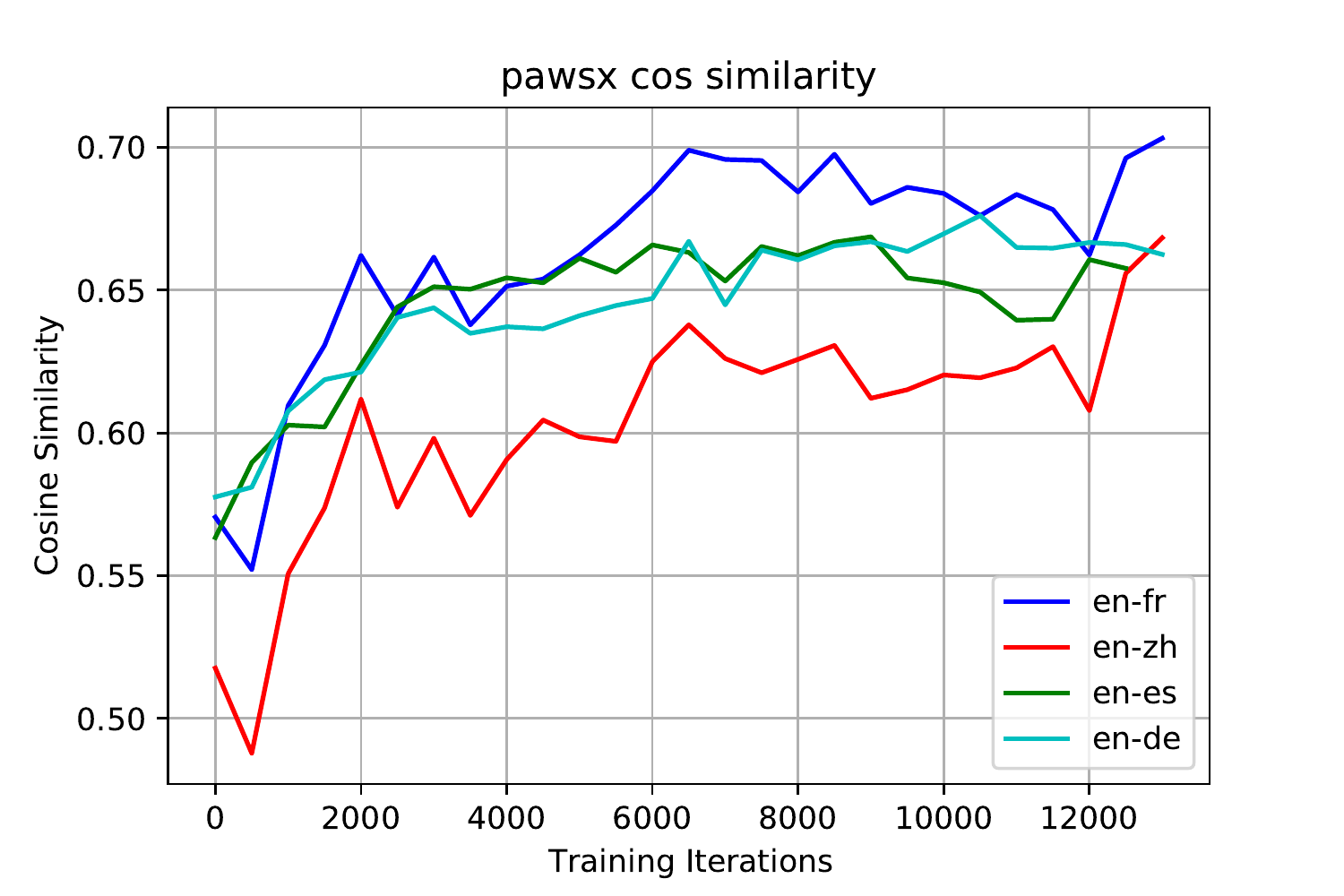}
    \caption{Average cosine similarity between parallel sentence representations (XLRS) in  pretrained mT5-base model fine-tuned on English PAWS-X data.}
    \label{pawsx-cos}
\end{figure}

 We plot the average cosine similarity between representations of parallel sentences (XLRS)\footnote{We use the mean-pooled token encoder hidden states as the sentence representation. We randomly sample 512 sentences from the test set of the MTG story completion task.} for each training iteration in two classification tasks: POS tagging and paraphrase identification (PAWS-X) at Figure \ref{pos-cos} and Figure \ref{pawsx-cos}, respectively.

In both tasks, the plot displays an increasing trend of XLRS between parallel sentences between English and all the other languages. Notably, languages that have the same script have a higher similarity. Our findings show that the fine-tuning process on classification tasks does make the sentence representations of different languages more similar. 

\begin{figure}
    \centering
    \includegraphics[width=180pt, height=110pt]{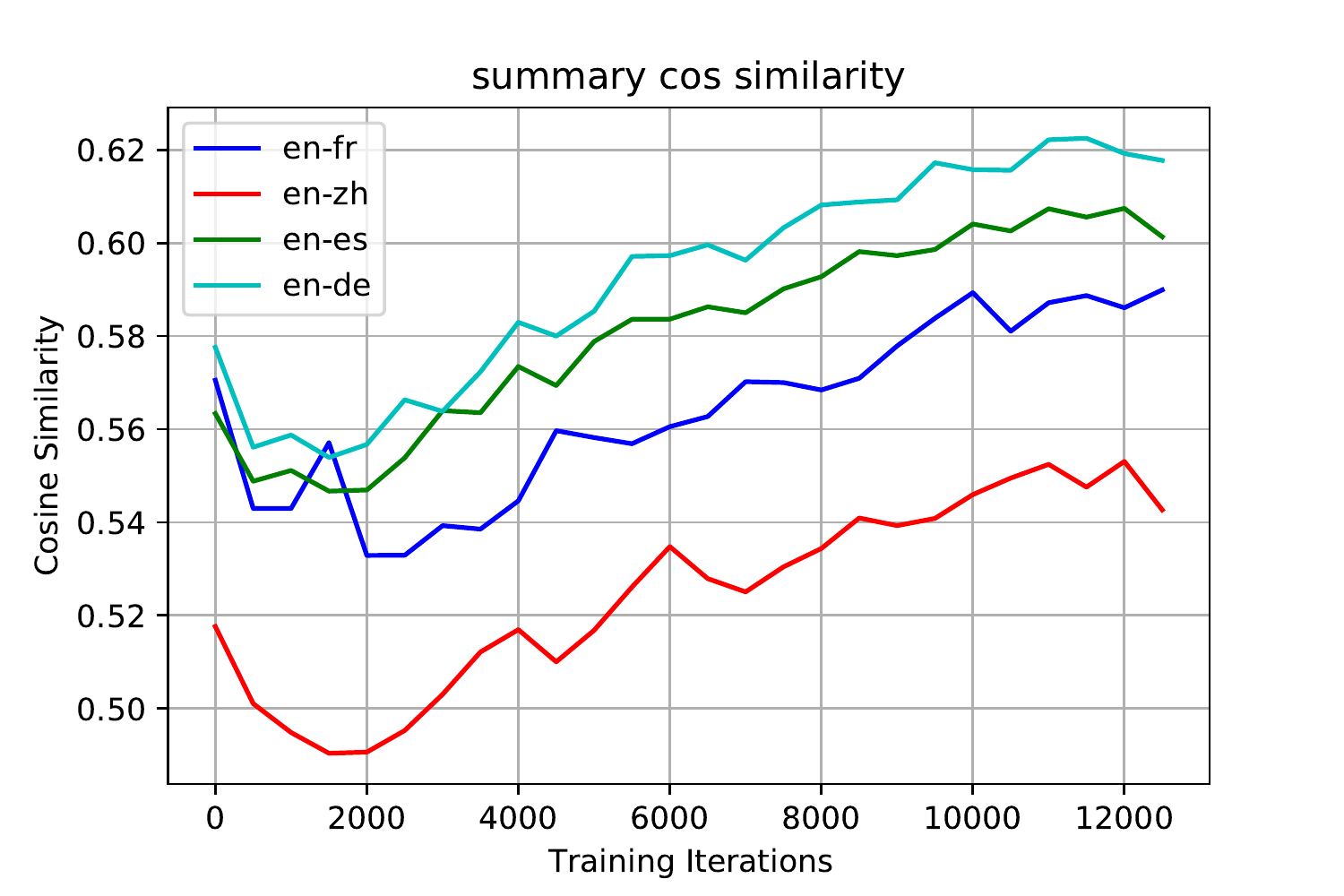}
    \caption{Average cosine similarity between parallel sentences representations (XLRS) of
    pretrained mT5-base model fine-tuned on English WikiLingua data.}
    \label{sum-cos}
\end{figure}

We then plotted the XLRS between representations of parallel sentences of a model when fine-tuned on WikiLingua: a summarization dataset in Figure \ref{sum-cos}. The average similarity gradually increases as we progress further into the training iterations, confirming our hypothesis that \textbf{fine-tuning on a single source language increases the XLRS between the source and other languages}.

Based on our findings, we conjecture
that the model jointly minimizes two metrics, resulting in cross-lingual transfer:
\begin{itemize}
    \item The Cross-Entropy loss between the predicted labels and the ground-truth labels, given an input in the source language (the standard training objective).
    \item The distance between parallel sentences of the source and target languages (increase in XLRS). 
\end{itemize}

And as a result, the cross-entropy loss between the predicted and ground-truth labels, given a context in the target language, is minimized, enabling cross-lingual transfer. 
\section{Unified View of Tasks}

\begin{figure*}[t]
    \centering
    \includegraphics[scale=0.8]{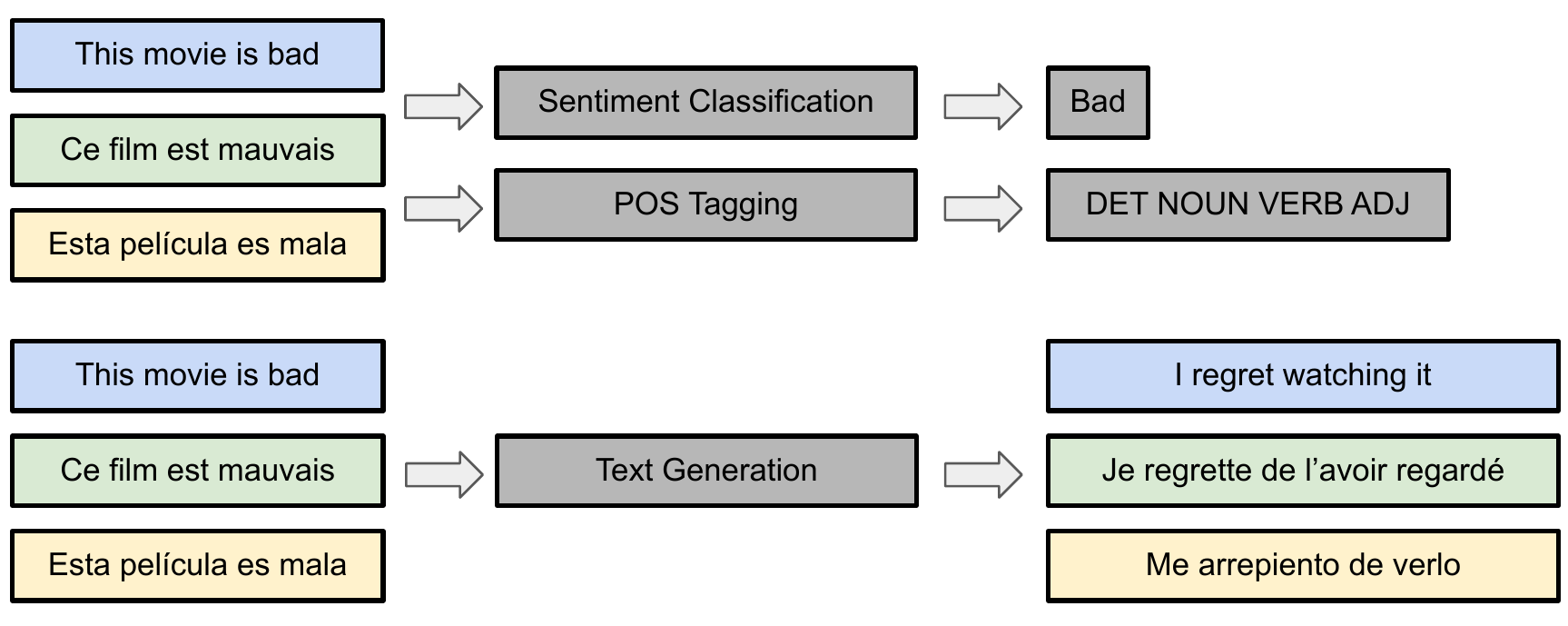}
    \caption{Illustration of the difference between classification tasks (top), where the model needs to map parallel sentences to the same label ($d=100\%$), and generation tasks (bottom), where the model needs to map parallel sentences into different labels ($d=0\%$). We use label overlap, $d$, to denote the fraction of parallel sentences that map to the same label.}
    \label{illustration}
\end{figure*}
With the intuition of how the model does cross-lingual transfer in classification tasks, we note that language generation is a classification on a large \emph{vocabulary set}, rather than a small label set. Thus, we point out that the reason why good performance with cross-lingual transfer on generation tasks is harder to achieve is actually caused by increasing XLRS. 

Figure \ref{illustration} illustrates our intuition. In classification tasks, the model needs to map parallel sentences to the same label. Ideally, the model produces identical representations for parallel sentences, resulting in the highest possible XLRS of 1. This is why model cross-lingual transfers better with a high XLRS. However, in generation tasks, if we view it as classifying over the entire vocabulary set, we are mapping parallel sentences to different labels. In an extreme case when XLRS is 1, the model fails to identify the source language, resulting in the common problem of the model producing an incorrect language \citep{xue-etal-2021-mt5}. We introduce the notion of \textbf{label overlap}, $d$, to measure the percentage of examples in a dataset where the model needs to map parallel sentences into the same label.

We use $\mathcal{C}$ to denote the set of all discrete contexts and $\mathcal{C}_s$ to denote the set of all discrete contexts in language $s$. In classification tasks, the model learns to predict the ground-truth label $\hat{y} = l(c)$ over a set of candidate labels $\mathcal{Y}$ for context $c \in \mathcal{C}$. Similarly, in a simplified view of generation, the model learns to predict the next word $v$ given context $c$. Therefore, we can essentially view language generation as classification where the label set is the entire vocabulary. In both cases, given context $c$, the model learns a probability distribution $p_{\cdot | c}$. The only difference is between their classification label set cardinality. 

We define cross-lingual label overlap as an indicator of difficulty to cross-lingual transfer for a given task at \hyperlink{section.5.1}{\S 5.1}. We then use a range of tasks: word-level classification (POS tagging \hyperlink{section.5.2}{\S 5.2}) - Sentence level classification (Entailment classification \hyperlink{section.5.3}{\S 5.3}) - Span Extraction (Extractive Generation \hyperlink{section.5.4}{\S 5.4}) - Summarization (Abstractive generation \hyperlink{section.5.5}{\S 5.5}) to show an increasing level of difficulty to perform cross-lingual transfer.

\subsection{Cross-lingual Label Overlap}
We denote a task's difficulty in transferring knowledge from one language to another by the percentage of overlap of their label set for parallel sentences. Given $n$ parallel sentences $\{c_s^1, c_s^2, ..., c_s^n\}$ and $\{c_t^1, c_t^2, ..., c_t^n\}$ in source language $s$ and target language $t$, the cross-lingual overlap $d$ for task $\alpha$ is defined as:
$$ d_{\alpha}(s, t) =  \frac{\sum_{i=1}^n \mathds{1}(l_{\alpha}(c_s^i) = l_{\alpha}(c_t^i))}{n}$$
In our analysis, we use English as the source language and evaluate the difficulty of performing cross-lingual transfer on other languages. The total label overlap for each task is the average label overlap for each target language.
$$ d_\alpha = \sum_{j=1}^m d_\alpha(\text{english}, t_j)$$
Where $t_j$ is the $j$th target language.

A higher $d_{\alpha}$ indicates an easier task to transfer knowledge from one language to another, whereas a lower $d_{\alpha}$ indicates a more difficult task to cross-lingual transfer.



\subsection{POS tagging}
\label{pos}
The word-level label overlap for part-of-speech tagging should be close to 100\%. With such a high percentage of label overlap, the model benefits from producing identical representations for parallel sentences to predict the same labels for different languages without supervision of the target language. 

For example, if the model maps the English sentence \textit{the dog ran} and the french sentence \textit{le chien courir} into nearly identical representations and simultaneously learns a function to map the English words to their respective POS tags "DET NOUN VERB", the model would also be able to predict the correct label for french even without supervision. We denote the amount of "label overlap" as a metric defining the difficulty for a model to perform cross-lingual transfer on it.

\subsection{Sentence Classification}
The classification task discussed in this section (PAWS-X) includes sentiment classification of a single sentence and entailment classification between two sentences. For semantically equivalent parallel sentences, their sentiment or entailment labels are always the same. Therefore, $d=100\%$. 

Ideally, in sentence classification tasks, parallel sentences in different languages should map to the same probability distribution. For example, if the English sentence \textit{I am happy} and the french sentence \textit{Je suis content} maps to nearly identical representations and the model learns to predict the sentiment in English, the model would be able to cross-lingual transfer the ability to predict sentiment in English to French without any supervision.


\subsection{Span Extraction}
Span extraction requires a model to select a correct answer span from a passage given a question. Even though the data in TyDiQA is in different languages, and not parallel, 16\% of the answer spans are pure numbers, and 50.6\% of answer spans are mainly composed of numbers (Time and Dates).

This indicates that span extraction is a harder task than sentence classification, but with such a high amount of label overlap, the task is solvable through cross-lingual transfer. 

\subsection{Generation}
The amount of label overlap in abstractive generation tasks (e.g. summarization, story completion, title generation) is close to zero as the model needs to predict words in completely different languages. The amount of label overlap for a subset of five languages (En, De, Fr, Es, Zh) of the WikiLingua \citep{ladhak-etal-2020-wikilingua} dataset is $d=0.13\%$\footnote{This small percentage represents mainly numbers and named entities (i.e. cities) that are the same across languages.}.

In a generation task, if the model maps the source and the target into identical representations, the model predicts the next word to be the same. Even if this is correct in semantics and possibly results in the code-switched results as shown in Figure \ref{examples}, the model fails to generate in the correct language.

\begin{table*}[]
\scriptsize
\begin{centering}
\begin{tabular}{@{}cllllllllllllllllll@{}}
\toprule
      & AR   & ZH   & CS   & NL   & EN   & FR   & HI   & ID   & IT   & JA   & KO   & PT   & RU   & ES   & TH   & TR   & VI   &  \\ \midrule
EN* & 17.4 & 15.1 & 17.8 & 20.1 & 39.6 & 22.4 & 9.1 & 23.0 & 20.3 & 14.6 & 17.3 & 23.8 & 15.3 & 23.3 & 17.9 & 17.5 & 21.9\\
EN    & 24.1 & 22.4 & 18.6 & 20.0 & 31.7 & 22.4 & 18.2 & 19.4 & 20.6 & \textbf{21.0} & 23.1 & 23.7 & 17.6 & 23.5 & 20.9 & 17.8 & 21.5 &  \\
EN+ZH & 24.3 & 27.5 & 20.4 & 22.6 & 33.2 & 23.8 & 18.8 & 21.6 & 22.1 & 18.4 & 21.2 & \textbf{27.2} & 20.2 & 24.9 & \textbf{21.8} & 18.2 & 24.3 &  \\ \midrule
DE    & 23.9 & 22.5 & 20.4 & 23.2 & 24.1 & 24.1  & 19.2 & 23.2 & 22.5 & 18.3 & 23.1 & 26.1 & 19.7 & 25.2 & 19.5 & 17.9 & 27.0 &  \\
DE+ZH & \textbf{24.8} & 27.9 & 21.2 & \textbf{24.2} & \textbf{26.0} & \textbf{25.2}  & 19.5 & \textbf{24.2} & \textbf{24.1} & 20.3 & \textbf{23.7} & 26.9 & \textbf{21.6} & \textbf{25.9} & \textbf{21.7} & 19.1 & 27.1 &  \\ \midrule
EN+DE & \textbf{24.9} & \textbf{25.2} & \textbf{21.4} & 24.0 & 22.3 & 25.3  & \textbf{20.1} & 23.7 & 22.8 & 19.3 & 24.1 & \textbf{27.2} & 20.3 & \textbf{25.8} & 20.0 & \textbf{19.3} & \textbf{27.4}   \\
\bottomrule
\end{tabular}
\end{centering}

\caption{ROUGE-L results on the WikiLingua \citep{ladhak-etal-2020-wikilingua} dataset. Left: Source languages that we fine-tune on. Top: Target language that we evaluate on. The bolded numbers refer to the highest zero-shot performance. Note that some of the directions are not zero-shot. The amount of training instances used in each row is the same. * indicate results reported in \citet{vu-2022-overcoming}.}
\label{wikilingua}
\end{table*}
\subsection{Analysis}
\begin{figure}
    \centering
    \includegraphics[width=180pt, height=110pt]{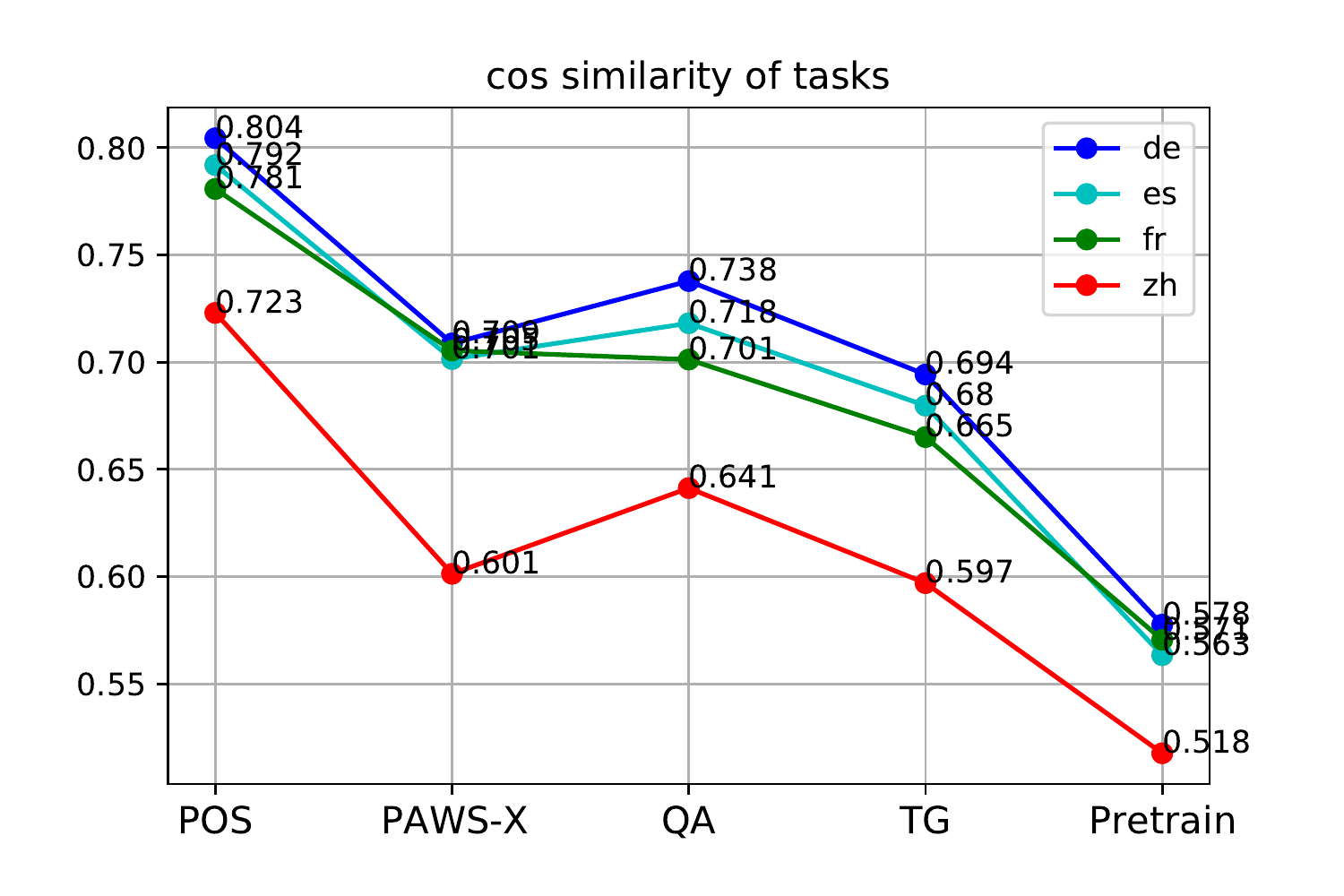}
    \caption{Average cosine similarity between representations of parallel sentences in English and 4 languages for the best performing model in 4 different tasks.}
    \label{seq-tasks-cos}
\end{figure}

We plotted the XLRS between English and four different languages in the best-performing English supervised models for the four tasks and the pretrained model at Figure \ref{seq-tasks-cos}.

The plot confirms our belief that for tasks (POS tagging, PAWS-X, TyDiQA) with large label overlap, the model cross-lingual transfers from increasing XLRS, whereas in generation tasks with label overlap close to zero (title generation), the best-performing model has a lower XLRS.

\begin{table}[]
\small
\centering
\begin{tabular}{@{}cccccc@{}}
\toprule
Task & POS  & PAWS-X & TG    & SG    & WikiLingua \\ \midrule
 $\rho$    & 0.89* & 0.91* & -0.37* & -0.39* & -0.33*      \\ \bottomrule
\end{tabular}
\caption{Spearman’s rank correlation $\rho$ between the average cosine similarity between parallel sentences in source and 4 target languages (De, Es, Fr, Zh) and the average zero-shot cross-lingual transfer performance (F1 for POS tagging, Acc. for PAWS-X and ROUGE-L for generation) on two classification tasks and three generation tasks, * indicates that the p-value is less than 0.05.}
\label{correlation}
\end{table}
Following \citet{yang2022enhancing}, we calculate the Spearman's rank correlation score between (a) XLRS between English and 4 target languages (German, French, Chinese, Spanish), and (b) The averaged zero-shot cross-lingual transfer performance in each task. The results are reported at Table \ref{correlation}. In both classification tasks, XLRS positively correlates with cross-lingual performance. In contrast, in our three generation tasks, XLRS negatively correlates with cross-lingual performance\footnote{We observed a stronger negative correlation between cosine similarity and ROUGE-2 in generation tasks but opted to report the ROUGE-L results to be consistent with our main results.}, indicating that \textbf{XLRS is strongly correlated to transfer performance in classification tasks but is detrimental to cross-lingual transfer in generation tasks.} 

\section{Text Generation Experiments}
\begin{figure}
    \centering
    \includegraphics[width=180pt, height=110pt]{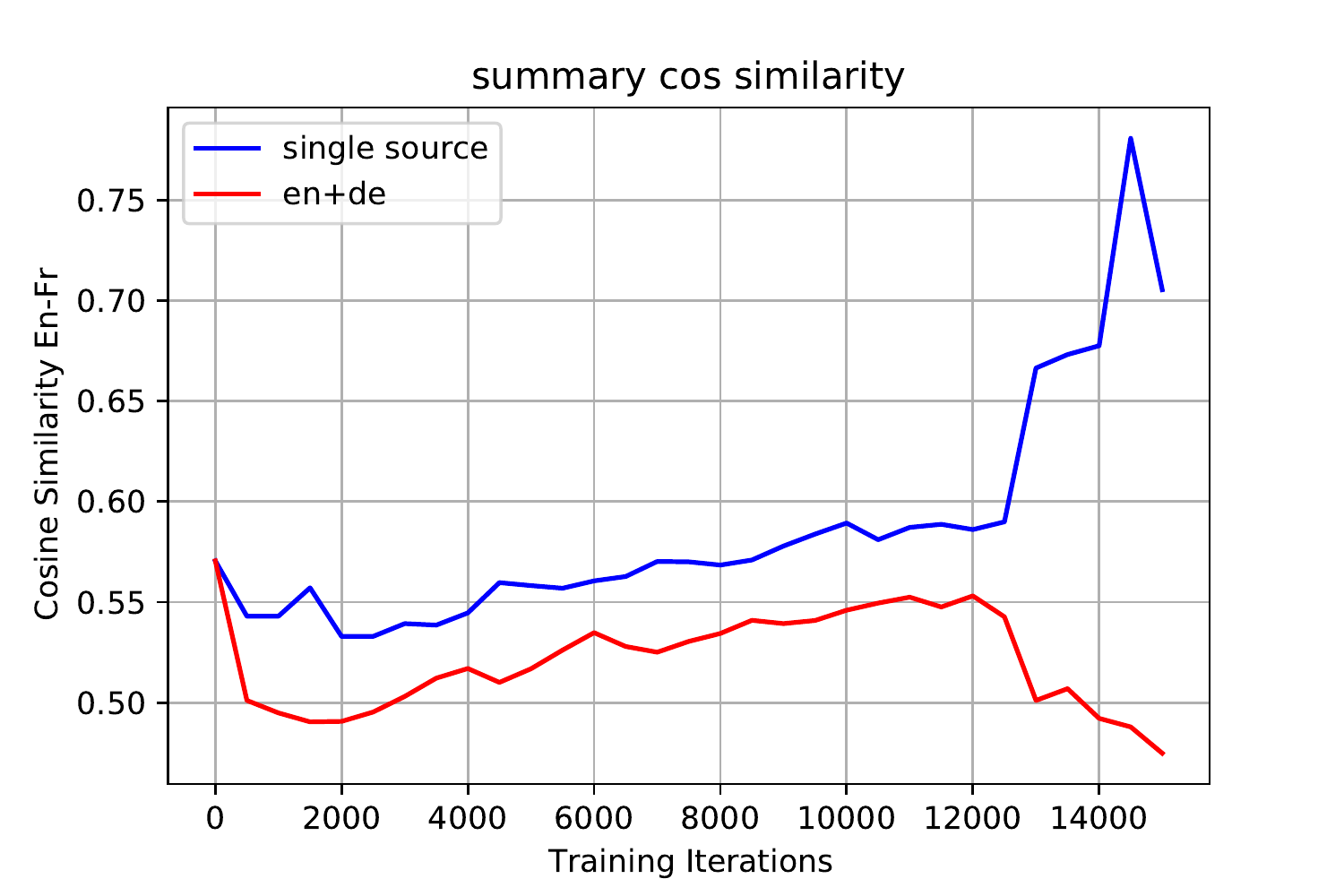}
    \caption{Average cosine similarity between representations of parallel sentences (XLRS) in English and French for model trained on one and two source languages.}
    \label{single_dual}
\end{figure}
Now that we know XLRS is negatively correlated with cross-lingual transfer in text generation, since calculating XLRS during every iteration is computationally expensive,
we wonder if we can \textbf{regularize XLRS implicitly}. Motivated by using auxiliary source languages improves machine translation \citep{xu-etal-2021-improving-multilingual} and few-shot cross-lingual transfer (\citealp{xu-murray-2022-por}; \citealp{dontstop-emnlp-2022}), we propose to use an additional source language to regularize XLRS.

To verify our hypothesis, we plot the XLRS between English and French during training on two different sources (En, De)in the story completion task, compared to training only on one source (En) in Figure \ref{single_dual}. We observe that when the model is only given one language as the source, the XLRS increases, whereas using two source languages allows the model to learn to control the XLRS from being too high, resulting in fewer accidental translations and better quality. 

To show that regularizing XLRS does result in better generation quality, we experiment with three semantically diverse generation tasks: Summarization 
 (\textbf{WikiLingua}), Title Generation (\textbf{TG}), and Story Completion (\textbf{SG}).

\subsection{Results}
Table \ref{wikilingua} shows the results of fine-tuning with multiple languages. We observe that adding Chinese to English data improves the performance in 13 out of 15 zero-shot directions\footnote{The results when the target language are Chinese (ZH) or English (EN) is not zero-shot.} compared to only using English. We point out that our improvement is not due to an increase in the amount of training data since we used the same amount of training data for all experiments. We further observe that adding Chinese as an additional language to German also improves the performance in all 14 zero-shot directions, which often results in the best zero-shot performance.

Table \ref{tg} and \ref{sg} show the ROUGE-L results in the title generation (TG) and story completion (SG) in the MTG \citep{chen-etal-2022-mtg} benchmark, respectively. Again, we are able to observe that using two source languages almost always improves the ROUGE-L score. Notably, using two related languages often results in degraded performance than using two unrelated languages with different scripts. We hypothesize that a language with a different script and order provides a more substantial regularization effect, preventing the cosine similarity between the source and target sentence representations from being too high.

\begin{table}[h]
\centering
\small
\begin{tabular}{@{}cllllll@{}}
\toprule
      & EN            & ES            & DE            & FR            & ZH            & Avg.            \\ \midrule
EN    & 32.3          & 26.0          & 24.4          & 25.3          & 19.6          & 25.5          \\
DE    & \textbf{30.2} & 24.7          & 22.5          & 23.9          & 18.7          & 24.0             \\
ZH    & 25.1          & 24.5          & 21.4          & 23.7          & 26.0          & 24.1        \\
DE+ZH & 29.2          & 24.8          & 23.6          & 24.9          & 22.7          & 25.0          \\
DE+EN & 27.8          & 24.6          & 23.6          & 22.3          & \textbf{20.8} & 23.8          \\
EN+ZH & 33.3          & \textbf{28.4} & \textbf{26.8} & \textbf{27.4} & 22.4          & \textbf{27.7} \\ \bottomrule
\end{tabular}
\caption{ROUGE-L results of title generation task in MTG benchmark. All experiments used same amount of data. The best zero-shot performance on each target language is bold.}
\label{tg}
\end{table}

\begin{table}[]
\centering
\small
\begin{tabular}{@{}cllllll@{}}
\toprule
\textbf{} & EN   & ES   & DE   & FR   & ZH   & Avg.   \\ \midrule
EN        & 29.1 & 28.9 & 27.8 & 28.9 & 20.3 & 27.0    \\
DE        & 29.3 & 28.7 & 29.9 & \textbf{32.0} & \textbf{20.4} & 28.0 \\
ZH        & 22.3 & 21.2 & 22.3 & 28.6 & 26.6 & 24.2  \\
DE+ZH     & \textbf{31.5} & \textbf{29.7} & 28.6 & 31.8 & 22.5 & 28.8 \\
DE+EN     & 30.4 & 27.3 & 26.4 & 28.2 & 19.8 & 26.4 \\
EN+ZH     & 31.8 & 28.5 & \textbf{29.3} & 29.1 & 28.6 & \textbf{29.5} \\ \bottomrule
\end{tabular}
\caption{ROUGE-L results of story completion task in MTG benchmark. All experiments used the same amount of data. The best zero-shot performance on each target language is bold.}
\label{sg}
\end{table}

To verify that our method helps against the accidental translation problem, we follow previous work \citep{vu-2022-overcoming} and calculate the language id confidence score on the source language and target language on the title generation task. The results are shown at Table \ref{lid} in Appendix A. Fine-tuning with multiple source languages helps the model learn which language it should produce.

\subsection{Model Selection Using Parallel Sentences}

Since XLRS negatively correlates with the performance of cross-lingual generation, we use it as a criterion for model selection in the absence of an annotated dev set. We report the performance on the WikiLingua dataset and the story completion task in MTG benchmark at Figure \ref{wikihow-criteria} and \ref{sg-selection}, when selecting the model using English dev set performance (\textbf{en-dev}), selecting the model with the lowest XLRS between English and the target language (\textbf{cos-sim}), and selecting the model using an annotated dev set on each target language (\textbf{tgt-dev}), which serves as an upper bound for true zero-shot cross-lingual transfer. 

In both tasks, selecting the model checkpoint  with the lowest XLRS results in better performance than using an English development set on all target languages. The performance is on average less than one ROUGE-L point less on Spanish and German in both datasets, compared to using an annotated dev set. Our method results in an average increase of 5 ROUGE-L points in a distant language (Chinese).\\
\begin{table}[h]
\centering
\begin{tabular}{@{}llllll@{}}
\toprule
        & ES   & DE   & FR   & ZH   & $\Delta$ \\ \midrule
en-dev  & 23.5 & 19.8 & 22.4 & 22.4 & -3.23 \\
cos-sim & 25.3 & 21.2 & 24.6 & 26.5 & -0.85 \\
tgt-dev & \textbf{25.4} & \textbf{21.9} & \textbf{25.4} & \textbf{28.3} & 0 \\ \bottomrule
\end{tabular}
\caption{ROUGE-L results by selecting model based on English development set (\textbf{en-dev}), similarity of representations between English and target language (\textbf{cos-sim}) and using target language development set (\textbf{tgt-dev}) on WikiLingua \citep{ladhak-etal-2020-wikilingua}.}
\label{wikihow-criteria}

\begin{tabular}{@{}llllll@{}}
\toprule
        & ES   & DE   & FR   & ZH  & $\Delta$ \\ \midrule
en-dev  & 28.9 & 27.8 & 28.9 & 20.3 &  -4.88 \\
cos-sim & 30.8 & 30.3 & \textbf{35.6} & 26.4 & -0.58 \\
tgt-dev & \textbf{31.2} & \textbf{30.5} & \textbf{35.6} & \textbf{28.1} & 0 \\ \bottomrule
\end{tabular}
\caption{ROUGE-L results by selecting model checkpoints in Story Completion (SG) task in MTG benchmark \citep{chen-etal-2022-mtg}.}
\label{sg-selection}
\end{table}


\section{Conclusion}
We show that multilingual LMs transfer supervision from one language to another by increasing Cross-Lingual Representation Similarity (XLRS). Such a learning process results in decent zero-shot cross-lingual transfer performance in  classification tasks but is harmful to text generation performance. We demonstrate that regularizing XLRS improves text generation quality and use parallel sentences to guide model selection without annotated data in the target languages. We believe that this is valuable under a practical setting \citep{artetxe-etal-2020-call} where we have access to parallel data between the source and target languages, but not task-specific data in the target language.

\section*{Limitations}
Our work sheds light on understanding the training dynamics of cross-lingual transfer learning of multilingual LMs. In our work, we selected to use English as the source of cross-lingual transfer following previous work \citep{vu-2022-overcoming}. We acknowledge that using other languages as the source language can provide benefits depending on the task (\citealp{lin-etal-2019-choosing}; \citealp{https://doi.org/10.48550/arxiv.2106.16171}). Our work does not focus on choosing source language to maximize downstream performance but instead focuses on the difference between classification tasks and generation tasks in cross-lingual transfer. 

Secondly, we acknowledge that some of the datasets (\citealp{yang-etal-2019-paws}; \citealp{chen-etal-2022-mtg}) used in our work is created by machine translation and human annotation. Previous studies have pointed out that translationese in datasets affects cross-lingual transfer performance (\citealp{artetxe-etal-2020-translation}; \citealp{artetxe-etal-2020-call}). We believe that translationese in datasets also has an impact on XLRS. We leave the study of how dataset features (size, quality, translationese) affect cross-lingual transfer for future work. 



\section*{Acknowledgements}
We sincerely thank Haoran Xu, 
Kelly Marchisio and Daniel Khashabi for their helpful suggestions.

\bibliography{anthology,custom}
\bibliographystyle{acl_natbib}
\newpage
\appendix
\section{Language Identification Scores}
\begin{table}[H]
\small
\centering
\begin{tabular}{@{}lllll@{}}
\toprule
                           & \multicolumn{2}{c}{FR} & \multicolumn{2}{c}{ES} \\ \midrule
\multicolumn{1}{l|}{}      & $\textrm{LID}_{DE}$         & $\textrm{LID}_{FR}$        & $\textrm{LID}_{DE}$         & $\textrm{LID}_{ES}$        \\
\multicolumn{1}{l|}{DE}    & 67.7       & 20.5      & 73.6       & 18.8      \\
\multicolumn{1}{l|}{DE+ZH} & 9.8        & 88.2      & 11.2       & 86.4      \\
\multicolumn{1}{l|}{DE+EN} & 15.2       & 76.2      & 14.9       & 78.4      \\ \bottomrule
\end{tabular}
\caption{Language identification confidence scores on the title generation task fine-tuned on single and multiple source languages.}
\label{lid}
\end{table}






\begin{table*}[ht]
\scriptsize
\begin{tabular}{@{}ll@{}}
\toprule
Task                                                               & Template                                                                                                                                                                                                                                                                                                                                                                                                                                                                                                                                                          \\ \midrule
\begin{tabular}[c]{@{}l@{}}Seq Tagging \\ (UDPOS)\end{tabular}     & \begin{tabular}[c]{@{}l@{}}Input: \textless{}extra\_id\_0\textgreater In \textless{}extra\_id\_2\textgreater my \textless{}extra\_id\_3\textgreater view \textless{}extra\_id\_4\textgreater it \textless{}extra\_id\_5\textgreater is \textless{}extra\_id\_6\textgreater significant\\ Output: \textless{}extra\_id\_0\textgreater ADP \textless{}extra\_id\_2\textgreater PRON \textless{}extra\_id\_3\textgreater NOUN \textless{}extra\_id\_4\textgreater PRON \textless{}extra\_id\_5\textgreater AUX  \textless{}extra\_id\_6\textgreater ADJ\end{tabular} \\ \midrule
\begin{tabular}[c]{@{}l@{}}Classification \\ (PAWS-X)\end{tabular} & \begin{tabular}[c]{@{}l@{}}Input: The original version was skipped in favor of the mild edition. \textless{}extra\_id\_0\textgreater  The mild version was skipped in favor of the original version.\\ Output: \textless{}extra\_id\_0\textgreater No.\end{tabular}                                                                                                                                                                                                                                                                                               \\ \midrule
\begin{tabular}[c]{@{}l@{}}QA\\  (TyDiQA)\end{tabular}             & \begin{tabular}[c]{@{}l@{}}Input: What is the surface area of the human cortex? \textless{}extra\_id\_0\textgreater\\ Output: \textless{}extra\_id\_0\textgreater 1.3 square feet\end{tabular}                                                                                                                                                                                                                                                                                                                                                                    \\ \midrule
\begin{tabular}[c]{@{}l@{}}Generation \\ (ByteCup)\end{tabular}    & \begin{tabular}[c]{@{}l@{}}Input: story: \{News article on Philadelphia Flower Show\} title: \textless{}extra\_id\_0\textgreater\\ Output: \textless{}extra\_id\_0\textgreater philly flower show will treat visitors to sights, sounds and scents of rainforest\end{tabular}                                                                                                                                                                                                                                                                                     \\ \bottomrule
\end{tabular}
\caption{\label{table1}Templates for casting tasks into a text-to-text format.}
\end{table*}

\end{document}